\documentclass[10pt,twocolumn,letterpaper]{article}
\usepackage[accsupp]{axessibility}  

\usepackage{iccv}
\usepackage{times}
\usepackage{epsfig}
\usepackage{graphicx}
\usepackage{amsmath}
\usepackage{amssymb}

\usepackage{graphicx}
\usepackage{amsmath}
\usepackage{amssymb}
\usepackage{booktabs}
\usepackage{multirow}
\usepackage{url}            
\usepackage{booktabs}       
\usepackage{amsfonts}       
\usepackage{nicefrac}       
\usepackage{microtype}      
\usepackage{xcolor}         
\usepackage{enumitem}
\usepackage{graphicx}
\usepackage{multirow}
\usepackage{amsmath}
\usepackage{amstext}
\usepackage{MnSymbol}
\usepackage{wasysym}
\usepackage{listings}
\usepackage{color,colortbl}
\usepackage{wrapfig}
\usepackage{soul}

\usepackage{bm}
\usepackage{algorithm}
\usepackage{algpseudocode}
\usepackage{textcomp}
\usepackage{multirow}

\definecolor{urlcolor}{rgb}{0.93,0.01,0.55}

\usepackage[breaklinks=true,bookmarks=false]{hyperref}

\iccvfinalcopy 

\def\iccvPaperID{6362} 

\ificcvfinal\fi

\begin{document}

\title{Rethinking Vision Transformers for MobileNet Size and Speed}

\author{Yanyu Li\\
Snap Inc., Northeastern University\\
{\tt\small li.yanyu@northeastern.edu}
\and
Ju Hu\\
Snap Inc.\\
{\tt\small jhu3@snap.com}
\and
Yang Wen\\
Snap Inc.\\
{\tt\small yangwenca@gmail.com}
\and
Georgios Evangelidis\\
Snap Inc.\\
{\tt\small gevangelidis@snap.com}
\and
Kamyar Salahi\\
UC Berkeley\\
{\tt\small kam.salahi@berkeley.edu}
\and
Yanzhi Wang\\
Northeastern University\\
{\tt\small yanz.wang@northeastern.edu}
\and
Sergey Tulyakov\\
Snap Inc.\\
{\tt\small stulyakov@snap.com}
\and
Jian Ren\\
Snap Inc.\\
{\tt\small jren@snap.com}
}


\maketitle
\ificcvfinal\thispagestyle{empty}\fi

\begin{abstract}
With the success of Vision Transformers (ViTs) in computer vision tasks, recent arts try to optimize the performance and complexity of ViTs to enable efficient deployment on mobile devices.
Multiple approaches are proposed to accelerate attention mechanism, improve inefficient designs, or incorporate mobile-friendly lightweight convolutions to form hybrid architectures.
However, ViT and its variants still have \emph{higher latency} or considerably \emph{more parameters} than lightweight CNNs, even true for the years-old MobileNet. 
In practice, latency and size are both crucial for efficient deployment on resource-constraint hardware. 
In this work, we investigate a central question, can transformer models run as fast as MobileNet and maintain a similar size? 
We revisit the design choices of ViTs and propose a novel supernet with low latency and high parameter efficiency. 
We further introduce a novel fine-grained joint search strategy for transformer models that can find efficient architectures by optimizing latency and number of parameters simultaneously. The proposed models, EfficientFormerV2, achieve $3.5\%$ higher top-1 accuracy than MobileNetV2 on ImageNet-1K with similar latency and parameters. This work demonstrate that properly designed and optimized vision transformers can achieve high performance even with MobileNet-level size and speed\footnote{Code: ~\href{https://github.com/snap-research/EfficientFormer}{\color{urlcolor}{https://github.com/snap-research/EfficientFormer}}.}.
\end{abstract}

\section{Introduction}

\begin{figure}[]
    \centering
    \includegraphics[width=1\linewidth]{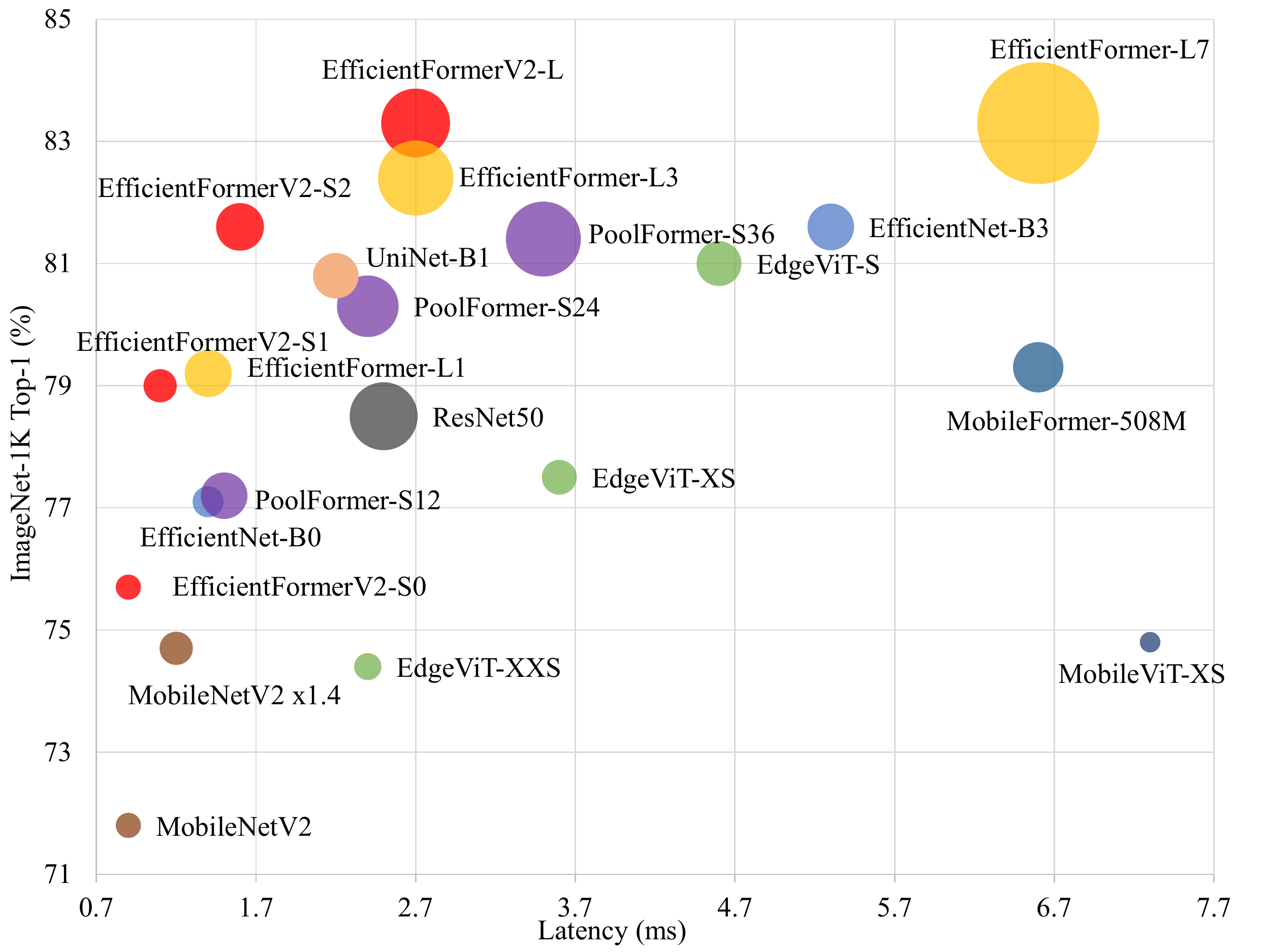}  
    \caption{\textbf{Comparison of model size, speed, and performance (top-1 accuracy on ImageNet-1K).}
    Latency is profiled by iPhone 12 (iOS 16). The area of each circle is proportional to the number of parameters (model size). EfficientFormerV2 achieves high performance with small model sizes and fast inference speed. }
    \label{fig:bubble}
\end{figure}
The promising performance of Vision Transformers (ViTs)~\cite{dosovitskiy2020vit} has inspired many follow-up works to further refine the model architecture and improve training strategies, leading to superior results on most computer vision benchmarks, such as classification~\cite{liu2021swin2,liu2021video,caron2021emerging,ma2022multimodal}, segmentation~\cite{xie2021segformer,cheng2021masked,cai2022efficientvit}, detection~\cite{carion2020end,li2021improved,si2022inception}, and image synthesis~\cite{esser2021taming,han2022show}.
As the essence of ViT, Multi Head Self Attention (MHSA) mechanism is proved to be effective in modeling spatial dependencies in $2$D images, enabling a global receptive field. 
In addition, MHSA learns second-order information with the attention heatmap as dynamic weights, which is a missing property in Convolution Neural Networks (CNNs) \cite{han2021connection}.
However, the cost of MSHA is also obvious--quadratic computation complexity with respect to the number of tokens (resolution).
Consequently, ViTs tend to be more computation intensive and have higher latency compared to widely adopted lightweight CNNs~\cite{howard2017mobilenets,howard2019searching}, especially on resource-constrained mobile devices, limiting their wide deployment in real-world applications.

Many research efforts~\cite{mehta2021mobilevit,mehta2022separable, pan2022edgevits,efficientformer} are taken to alleviate this limitation. 
Among them, one direction is to reduce the quadratic computation complexity of the attention mechanism.
Swin \cite{liu2021swin} and following works~\cite{dong2022cswin,liu2021swin2} propose window-based attention such that the receptive field is constrained to a pre-defined window size, which also inspires subsequent work to refine attention patterns \cite{chen2021crossvit,wang2021crossformer,wu2022pale,pan2022fast}. 
With the pre-defined span of attention, the computation complexity becomes linear to resolution. 
However, sophisticated attention patterns are generally difficult to support or accelerate on mobile devices because of intensive shape and index operations. 
Another track is to combine lightweight CNN and attention mechanism to form a hybrid architecture \cite{mehta2021mobilevit,chen2021mobile,maaz2022edgenext}. 
The benefit comes two-fold. First, convolutions are shift invariant and are good at capturing local and detailed information, which can be considered as a good complement to ViTs \cite{han2021connection}. 
Second, by placing convolutions in the early stages while placing MHSA in the last several stages to model global dependency, we can naturally avoid performing MHSA on high resolution and save computations \cite{liu2022uninet}. 
Albeit achieving satisfactory performance, the latency and model size are still less competitive compared to lightweight CNNs. 
For instance, MobileViT~\cite{mehta2021mobilevit} achieves better performance than MobileNetV2 while being at least $5\times$ slower on iPhone 12. 
As applicable to CNNs, architecture search, pruning, and quantization techniques are also thoroughly investigated \cite{jin2021teachers,jin2022f8net,liu2021post,kim2021bert,chavan2022vision,efficientformer,liu2022uninet}. 
Nevertheless, these models still emerge obvious weaknesses,   
\emph{e.g.}, EfficientFormer-L1~\cite{efficientformer} achieves comparable speed and better performance than MobileNetV2$\times1.4$, while being $2\times$ larger. 
Thus, a \emph{practical} yet \emph{challenging} question arises:

\emph{Can we design a transformer-based model that is both light and fast, and preserves high performance}?

In this work, we address the above question and propose a new family of mobile vision backbones. 
We consider three vital factors: \emph{number of parameters}, \emph{latency}, and \emph{model performance}, as they reflect disk storage and mobile applications.
First, we introduce \emph{novel architectural improvements} to form a strong design paradigm.  
Second, we propose a \emph{fine-grained} architecture search algorithm that jointly optimizes model size and speed for transformer models. 
With our network design and search method, we obtain a series of models under various constraints of model size and speed while maintaining high performance, named EfficientFormerV2. 
Under the exact same size and latency (on iPhone 12), EfficientFormerV2-S0 outperforms MobileNetV2 by $3.5\%$ higher top-1 accuracy on ImageNet-1K~\cite{deng2009imagenet}.
Compared to EfficientFormer-L1~\cite{efficientformer}, EfficientFormerV2-S1 has similar performance while being $2\times$ smaller and $1.3\times$ faster (Tab.~\ref{tab:comparison}). 
We further demonstrate promising results in downstream tasks such as detection and segmentation (Tab.~\ref{tab:coco}).
Our contributions can be concluded as follows.
\begin{itemize}[leftmargin=1em]
\setlength
\itemsep{-0.25em}
    \item We comprehensively study mobile-friendly design choices and introduce novel changes, which is a practical guide to obtaining ultra-efficient vision transformer backbones.
    \item We propose a novel fine-grained joint search algorithm that simultaneously optimizes model size and speed for transformer models, achieving superior Pareto optimality.
    \item For the first time, we show that vision transformer models can be as small and fast as MobileNetV2 while obtaining much better performance. EfficientFormerV2 can serve as a strong backbone in various downstream tasks.
\end{itemize}

\section{Related Work}
Vaswani \textit{et al.}~\cite{vaswani2017attention} propose attention mechanism to model sequences in NLP task, which forms transformer architecture. 
Transformers are later adopted to vision tasks by Dosovitskiy \textit{et al.}~\cite{dosovitskiy2020vit} and Carion~\textit{et al.}~\cite{carion2020end}. 
DeiT~\cite{touvron2021training} improves ViT by training with distillation and achieves competitive performance against CNNs. 
Later research further improves ViTs by incorporating hierarchical design~\cite{wang2021pyramid,touvron2021going}, injecting locality with the aid of convolutions~\cite{guo2021cmt,dai2021coatnet,han2021connection,si2022inception}, or exploring different types of token mixing such as local attention~\cite{liu2021swin,dong2022cswin}, spatial MLP mixer~\cite{touvron2021resmlp,tolstikhin2021mixer}, and non-parameterized pool mixer~\cite{yu2021metaformer}. 
With appropriate changes, ViTs demonstrate strong performance in downstream vision tasks~\cite{xie2021segformer,zhang2022nested,zhang2022topformer,lee2021vision,lee2021vitgan,esser2021taming,zeng2021improving}. 
To benefit from the advantageous performance, efficient deployment of ViTs has become a research hotspot, especially for mobile devices~\cite{mehta2021mobilevit,chen2021mobile,pan2022edgevits,maaz2022edgenext}.
For reducing the computation complexity of ViTs,  
many works propose new modules and architecture design~\cite{Nikita2020,hassani2021escaping,fayyaz2021ats,Wei2022,Renggli2022}, while others eliminate redundancies in attention mechanism~\cite{wang2021crossformer,heo2021rethinking,chen2021regionvit,li2021localvit,chu2021twins,rao2021dynamicvit,Zhengzhong2022,hydraattention}. 
Similar to CNNs, architecture search~\cite{chen2021autoformer,gong2022nasvit,chavan2022vision,zhou2022training,liu2022uninet,fbnetv3,fbnetv5}, pruning~\cite{zhang2022platon}, and quantization~\cite{liu2021post} are also explored for ViTs. 

We conclude two major drawbacks of the study in efficient ViT.
First, many optimizations are not suitable for mobile deployment. For example, the quadratic computation complexity of the attention mechanism can be reduced to linear by regularizing the span or pattern of attention mechanism~\cite{liu2021swin,dong2022cswin,chen2021crossvit,deformableattention}. Still, the sophisticated reshaping and indexing operations are not even supported on resource-constrained devices~\cite{efficientformer}. 
It is crucial to rethink the mobile-friendly designs. 
Second, though recent hybrid designs and network search methods reveal efficient ViTs with strong performance~\cite{mehta2021mobilevit,liu2022uninet,efficientformer}, they mainly optimize the Pareto cure for one metric while being less competitive in others. 
For example, MobileViT~\cite{mehta2021mobilevit} is parameter efficient while being times slower than lightweight CNNs~\cite{sandler2018mobilenetv2,tan2019efficientnet}. 
EfficientFormer~\cite{efficientformer} wields ultra-fast speed on mobile, but the model size is enormous. 
LeViT~\cite{levit} and MobileFormer~\cite{chen2021mobile} achieve favorable FLOPs at the cost of redundant parameters. 

\section{Rethinking Hybrid Transformer Network} \label{sec:3rethink}

\begin{figure*}[]
    \centering
    \includegraphics[width=1\linewidth]{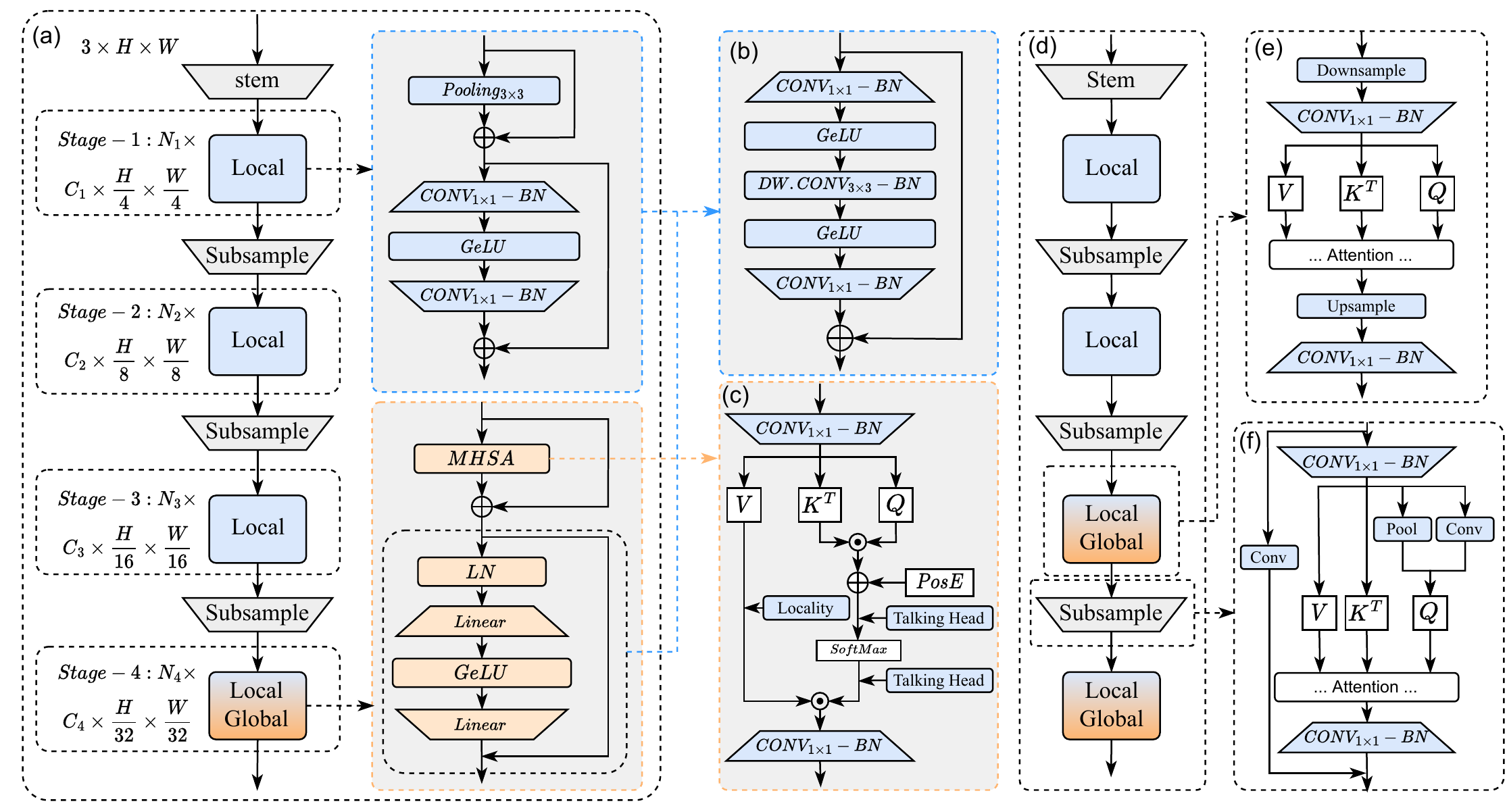}  
    \caption{\textbf{Network architectures.} We consider three metrics, \emph{i.e.}, model performance, size, and inference speed, and study the models that improve any metric without hurting others. (a) Network of EfficientFormer~\cite{efficientformer} that serves as a baseline model. (b) Unified FFN (Sec.~\ref{sec:mixer}). (c) MHSA improvements (Sec.~\ref{sec:improve_mhsa}). (d)\&(e) Attention on higher resolution (Sec.~\ref{sec:higher_res}). (f) Dual-Path Attention downsampling (Sec.~\ref{sec:subsample}).}
    \label{fig:main}
\end{figure*}

\begin{table*}[]
\small
\centering
\caption{\textbf{Number of parameters, latency, and performance} for various design choices. The latency is tested on iPhone 12. Top-1 accuracy is obtained by validating models on ImageNet-1K for the classification task.}\label{tab:rethink_design}
\resizebox{1.0\linewidth}{!}{
\small
\begin{tabular}{cccccc}
\toprule
Section                & Method                         & \#Params (M) & MACs (G) & Latency (ms) & Top-1 (\%) \\
\hline
(Baseline)             & EfficientFormer-L1             & 12.25        & 1.30     & 1.4          & 79.2     \\
\hline
\multirow{2}{*}{Sec.~\ref{sec:mixer}} &  Pool Mixer $\rightarrow$ DWCONV$_{3\times3}$                    & 12.27        & 1.30     & 1.4          & 79.8     \\
                      & $\checkmark$ Feed Forward Network 
                      & 12.37        & 1.33     & 1.4          & 80.3     \\
                      \hline
\multirow{2}{*}{Sec.~\ref{sec:search}} & $\checkmark$ Vary Depth and Width               & 12.24        & 1.20     & 1.3          & 80.5     \\
                      & 5-Stage Network                        & 12.63        & 1.08     & 1.5          & 80.3     \\
                      \hline
Sec.~\ref{sec:improve_mhsa}                  & $\checkmark$ Locality in $V$ \& Talking Head  & 12.25        & 1.21     & 1.3          & 80.8     \\
\hline
\multirow{2}{*}{Sec.~\ref{sec:higher_res}} & Attention at Higher Resolution & 13.10        & 1.48     & 3.5          & 81.7     \\
                      & $\checkmark$ Stride Attention              & 13.10        & 1.31     & 1.5          & 81.5     \\
                      \hline
\multirow{2}{*}{Sec.~\ref{sec:subsample}}                 &  Attention Downsampling        & 13.18        & 1.33     & 1.6          & 81.4     \\
 & $\checkmark$ Dual-Path Attention Downsampling        & 13.40        & 1.35     & 1.6          & 81.8     \\
\bottomrule
\end{tabular}
}
\end{table*}

In this section, we study the design choices for efficient ViTs and introduce the changes that lead to the smaller size and faster speed without a performance drop.  EfficientFormer-L1~\cite{efficientformer} is used as a baseline model given its superior performance on mobile devices. 

\subsection{Token Mixers \emph{vs.} Feed Forward Network}\label{sec:mixer}
Incorporating local information can improve performance and make ViTs more robust to the absence of explicit positional embedding~\cite{cai2022efficientvit}. 
PoolFormer~\cite{yu2021metaformer} and EfficientFormer~\cite{efficientformer} employ $3\times3$ average pooling layers (Fig.~\ref{fig:main}{(a)}) as local token mixer. Replacing these layers with depth-wise convolutions (DWCONV) of the same kernel size does not introduce latency overhead, while the performance is improved by $0.6\%$ with negligible extra parameters ($0.02$M).
Further, recent work \cite{gong2022nasvit,cai2022efficientvit} suggest that it is also beneficial to inject local information modeling layers in the Feed Forward Network (FFN) in ViTs to boost performance with minor overhead. 
It is noteworthy that by placing extra depth wise $3\times 3$ convolutions in FFNs to capture local information, the functionality of original local mixer (pooling or convolution) is duplicated.
Based on these observations, we remove the explicit residual-connected local token mixer and move the dept-wise $3\times3$ CONV into the FFN, to get a unified FFN (Fig.~\ref{fig:main}{(b)}) with locality enabled. 
We apply the unified FFN to all stages of the network, as in Fig.~\ref{fig:main}{(a,b)}. 
Such design modification simplifies the network architecture to only two types of blocks (local FFN and global attention), and boosts the accuracy to $80.3\%$ at the same latency (see Tab.~\ref{tab:rethink_design}) with minor overhead in parameters ($0.1$M). 
More importantly, this modification allows us to \emph{directly search the network depth} with the exact number of modules in order to extract local and global information, 
especially at the late stages of the network, as discussed in Sec.~\ref{sec:joint_search}.



\subsection{Search Space Refinement}\label{sec:search}
With the unified FFN and the deletion of residual-connected token mixer, we examine whether the search space from EfficientFormer is still sufficient, especially in terms of depth. 
We vary the network depth (number of blocks in each stage) and width (number of channels), and find that deeper and narrower network leads to  better accuracy ($0.2\%$ improvement), less parameters ($0.13$M reduction), and lower latency ($0.1$ms acceleration), as in Tab.~\ref{tab:rethink_design}. 
Therefore, we set this network as a new baseline (accuracy $80.5\%$) to validate subsequent design modifications, and enable a deeper supernet for architecture search in Sec.~\ref{sec:joint_search}. 

In addition, 5-stage models with further down-sized spatial resolution ($\frac{1}{64}$) have been widely employed in efficient ViT arts~\cite{levit,chen2021mobile,liu2022uninet}. 
To justify whether we should search from a 5-stage supernet, we append an extra stage to current baseline network and verify the performance gain and overhead. 
It is noteworthy that though computation overhead is not a concern given the small feature resolution, the additional stage is parameter intensive. 
As a result, we need to shrink the network dimension (depth or width) to align parameters and latency to the baseline model for fair comparison. 
As seen in Tab.~\ref{tab:rethink_design}, the best performance of the 5-stage model surprisingly drops to $80.31\%$ with more parameters ($0.39$M) and latency overhead ($0.2$ms), despite the saving in MACs ($0.12$G). 
This aligns with our intuition that the fifth stage is computation efficient but parameter intensive.
Given that 5-stage network can not introduce more potentials in our size and speed scope, we stick to 4-stage design. 
This analysis also explains why some ViTs offer an excellent Pareto curve in MACs-Accuracy, but tend to be quite redundant in size~\cite{levit,chen2021mobile}. 
As the most important takeaway, optimizing single metric is easily trapped, and the proposed joint search in Sec.~\ref{sec:joint_search} provides a feasible solution to this issue.

\subsection{MHSA Improvements}\label{sec:improve_mhsa}
We then study the techniques to improve the performance of attention modules without raising extra overhead in model size and latency. 
As shown in Fig.~\ref{fig:main}{(c)}, we investigate two approaches for MHSA.
First, we inject local information into the Value matrix ($V$) by adding a depth-wise $3\times3$ CONV, which is also employed by \cite{gong2022nasvit,si2022inception}. Second, we enable communications between attention heads by adding fully connected layers across head dimensions \cite{shazeer2020talking} that are shown as Talking Head in Fig.~\ref{fig:main}{(c)}. 
With these modifications, we further boost the performance to $80.8\%$ with similar parameters and latency compared to the baseline model.

\subsection{Attention on Higher Resolution}\label{sec:higher_res}
Attention mechanism is beneficial to performance. However, applying it to high-resolution features harms mobile efficiency since it has quadratic time complexity corresponding to spatial resolution.
We investigate strategies to efficiently apply MHSA to higher resolution (early stages).
Recall that in the current baseline network obtained in Sec.~\ref{sec:improve_mhsa}, MHSA is only employed in the last stage with $\frac{1}{32}$ spatial resolution of the input images. 
We apply extra MHSA to the second last stage with $\frac{1}{16}$ feature size, and observe $0.9\%$ gain in accuracy. 
On the down side, the inference speed slows down by almost $2.7\times$. Thus, it is necessary to properly reduce complexity of the attention modules. 

Although some work propose window-based attention~\cite{liu2021swin,dong2022cswin}, or downsampled Keys and Values \cite{li2022next} to alleviate this problem, we find that they are not best-suited options for mobile deployment. Window-based attention is difficult to accelerate on mobile devices due to the sophisticated window partitioning and reordering. As for downsampling Keys ($K$) and Values ($V$) in \cite{li2022next}, full resolution Queries ($Q$) are required to preserve the output resolution ($\mathbf{Out}$) after attention matrix multiplication:
\begin{equation}
\small
    \mathbf{Out}_{[B, H, N, C]} = ( Q_{[B, H, N, C]} \cdot K^T_{[B, H, C, \frac{N}{2}]} ) \cdot V_{[B, H, \frac{N}{2}, C]},
\end{equation}
where $B$, $H$, $N$, $C$ denotes batch size, number of heads, number of tokens, and channel dimension respectively. 
Based on our test, the latency of the model merely drops to $2.8$ms, which is still $2\times$ slower than the baseline model.

Therefore, to perform MHSA at the earlier stages of the network, we downsample all Query, Key, and Value to a fixed spatial resolution ($\frac{1}{32}$) and interpolate the outputs from the attention back to the original resolution to feed into the next layer, as shown in Fig.~\ref{fig:main}({(d)\&(e)}).
We refer to this method as Stride Attention. 
As in Tab.~\ref{tab:rethink_design}, this simple approximation significantly reduces the latency from $3.5$ms to $1.5$ms and preserves a competitive accuracy ($81.5\%$ \emph{vs.} $81.7\%$).  

\subsection{Dual-Path Attention Downsampling}\label{sec:subsample}
Most vision backbones utilize strided convolutions or pooling layers to perform a static and local downsampling and form a hierarchical structure. 
Some recent works start to explore attention downsampling. For instance, LeViT~\cite{levit} and UniNet~\cite{liu2022uninet} propose to halve feature resolution via attention mechanism to enable context-aware downsampling with the global receptive field. Specifically, the number of tokens in Query is reduced by half so that the output from the attention module is downsampled:
\begin{equation}
\small
    \mathbf{Out}_{[B, H, \frac{N}{2}, C]} = ( Q_{[B, H, \frac{N}{2}, C]} \cdot K^T_{[B, H, C, N]} ) \cdot V_{[B, H, N, C]}.
\end{equation}
However, it is nontrivial to decide how to reduce the number of tokens in Query. 
Graham \emph{et al.} empirically use pooling to downsample Query~\cite{levit}, while Liu \emph{et al.} propose to search for local or global approaches~\cite{liu2022uninet}. 
To achieve acceptable inference speed on mobile devices, applying attention downsampling to early stages with high resolution is not favorable, restricting the values of existing works that search different downsampling approaches at higher-resolution.

Instead, we propose a combined strategy, which is dual-path attention downsampling, that wields both locality and global dependency, as in Fig.~\ref{fig:main}{(f)}. 
To get downsampled Queries, we use pooling as static local downsampling,  $3\times 3$ DWCONV as learnable local downsampling, and combine and project the results into Query dimension. In addition, the attention downsampling module is residual connected to a regular strided CONV to form a local-global manner, similar to the downsampling bottlenecks~\cite{he2016deep} or inverted bottlenecks~\cite{sandler2018mobilenetv2}.
As shown in Tab.~\ref{tab:rethink_design}, with slightly more parameters and latency overhead, we further improve the accuracy to $81.8\%$ with dual-path attention downsampling, which also has better performance than only using attention module for subsampling, \emph{i.e.}, attention downsampling.

\section{EfficientFormerV2}
As discussed, current arts merely focus on optimizing one metric, thus are either redundant in size~\cite{efficientformer} or slow in inference~\cite{mehta2021mobilevit}. 
To find the most suitable vision backbones for mobile deployment, we propose to jointly optimize model size and speed. 
Furthermore, the network designs in Sec.~\ref{sec:3rethink} 
favor a deeper network architecture (Sec.~\ref{sec:search}) and more attentions (Sec.~\ref{sec:higher_res}), calling for an improved search space and algorithm. 
In what follows, we present the supernet design of EfficientFormerV2 and its search algorithm. 

\subsection{Design of EfficientFormerV2}
As discussed in Sec.~\ref{sec:search}, we employ a 4-stage hierarchical design which obtains feature sizes in  $\{\frac{1}{4},\frac{1}{8},\frac{1}{16},\frac{1}{32}\}$ of the input resolution. 
Similar to its predecessor~\cite{efficientformer}, EfficientFormerV2 starts with a small kernel convolution \texttt{stem} to embed input image instead of using inefficient 
embedding of non-overlapping patches,
\begin{equation}
\small
    \mathbb{X}_{i|_{i=1}, j|_{j=1}}^{B,C_{j|_{j=1}},\frac{H}{4},\frac{W}{4}} = \texttt{stem}(\mathbb{X}_0^{B,3,H,W}),
\end{equation}
where $B$ denotes the batch size, $C$ refers to channel dimension (also represents the width of the network), $H$ and $W$ are the height and width of the feature, $\mathbb{X}_j$ is the feature in stage $j$, $j \in \{1,2,3,4\}$, and $i$ indicates the $i$-th layer. 
The first two stages capture local information on high resolutions; thus we only employ the unified FFN (\texttt{FFN}, Fig.~\ref{fig:main}{(b)}), 
\begin{equation}
\small
        \mathbb{X}_{i+1, j}^{B,C_j,\frac{H}{2^{j+1}},\frac{W}{2^{j+1}}} = \texttt{S}_{i,j} \cdot \texttt{FFN}^{C_j,E_{i,j}}(\mathbb{X}_{i,j}) + \mathbb{X}_{i,j},
        \label{eq:local}
\end{equation}
where $\texttt{S}_{i,j}$ is a learnable layer scale \cite{yu2021metaformer} and the $\texttt{FFN}$ is constructed by two properties: stage width $C_j$ and a per-block expansion ratio $E_{i,j}$.
Note that each \texttt{FFN} is residual connected. 
In the last two stages, both local \texttt{FFN} and global \texttt{MHSA} blocks are used. Therefore,  on top of Eqn.~\ref{eq:local}, global blocks are defined as:
\begin{equation}
\small
        \mathbb{X}_{i+1,j}^{B,C_j,\frac{H}{2^{j+1}},\frac{W}{2^{j+1}}} = \texttt{S}_{i,j} \cdot \texttt{MHSA}(\texttt{Proj}(\mathbb{X}_{i,j})) + \mathbb{X}_{i,j},
        \label{eq:global}
\end{equation}
where Queries ($Q$), Keys ($K$), and Values ($V$) are projected from input features through linear layers $Q, K, V \leftarrow \texttt{Proj}(\mathbb{X}_{i,j})$, and 
\begin{equation}
\small
  \texttt{MHSA}(Q,K,V) = \texttt{Softmax}(Q\cdot K^T + \texttt{ab}) \cdot V,
\end{equation}
with \texttt{ab} as a learnable attention bias for position encoding.

\subsection{Jointly Optimizing Model Size and Speed} \label{sec:joint_search}
Though the baseline network EfficientFormer~\cite{efficientformer} is found by latency-driven search and wields fast inference speed on mobile, there are two major drawbacks for the search algorithm. 
First, the search process is merely constrained by speed, resulting in the final models being parameter redundant, as in Fig.~\ref{fig:bubble}.
Second, it only searches for depth (number of blocks $N_j$ per stage) and stage width $C_j$, which is in a \emph{coarse-grained} manner. In fact, the majority of computations and parameters of the network are in $\texttt{FFN}$s, and the parameter and computation complexity are linearly related to its expansion ratio $E_{i,j}$. $E_{i,j}$ can be specified independently for each $\texttt{FFN}$ without the necessity to be identical. Thus, searching $E_{i,j}$ enables a more \emph{fine-grained} search space where the computations and parameters can distribute \emph{flexibly} and \emph{non-uniformly} within each stage. 
This is a missing property in most recent ViT NAS arts \cite{gong2022nasvit,liu2022uninet,efficientformer}, where $E_{i,j}$ remains identical per stage. 
We propose a search algorithm that enables a flexible per-block configuration, with joint constraints on size and speed, and finds vision backbones best suited for mobile devices.  

\subsubsection{Search Objective} \label{sec:objective}
First, we introduce the metric guiding our joint search algorithm. Given the fact that the size and latency of a network all matter when evaluating mobile-friendly models, we consider a generic and fair metric that better understands the performance of a network on mobile devices.
Without loss of generality, we define a \underline{M}obile \underline{E}fficiency \underline{S}core (\texttt{MES}):
\begin{equation}
\small
    \texttt{MES} = Score \cdot \prod_i (\frac{M_i}{U_i})^{-\alpha_i},
    \label{eq:mescore}
\end{equation}
where $i \in \{ {size}, latency, ...\}$ and $\alpha_i \in (0, 1]$ indicating the corresponding importance. $M_i$, and $U_i$ represent the metric and its unit.
$Score$ is a pre-defined base score set as $100$ for simplicity.
Model size is calculated by the number of parameters, and latency is measured as running time when deploying models on devices.
Since we focus on mobile deployment, the size and speed of MobileNetV2 are used as the unit.
Specifically, we define $U_{size}=3\text{M}$, and $U_{latency}$ as $1$ms latency on iPhone 12 (iOS 16) deployed with CoreMLTools~\cite{coreml2021}. To emphasize speed,
we set $\alpha_{latency} = 1.0$ and $\alpha_{size} = 0.5$.
Decreasing size and latency
can lead to a higher \texttt{MES}, and we search for Pareto optimality on \texttt{MES}-Accuracy.
The form of \texttt{MES} is general and can be extended to other metrics of interest, such as inference-time memory footprint and energy consumption. Furthermore, the importance of each metric is easily adjustable by appropriately defining $\alpha_i$.

\subsubsection{Search Space and SuperNet}
\noindent\textbf{Search space} consists of: (i) the depth of the network, measured by the number of blocks $N_j$ per stage, (ii) the width of the network, \emph{i.e.}, the channel dimension $C_j$ per stage, and (iii) expansion ratio $E_{i,j}$ of each \texttt{FFN}. 
The amount of \texttt{MHSA} can be seamlessly determined during depth search, which controls the preservation or deletion of a block in the supernet. Thus, we set every block as \texttt{MHSA} followed by \texttt{FFN} in the last two stages of the supernet and obtain subnetworks with the desired number of global \texttt{MHSA} by depth search. 

\noindent\textbf{Supernet} is constructed by using a slimmable network~\cite{slimmable} that executes at elastic depth and width to enable a pure evaluation-based search algorithm. 
Elastic depth can be naturally implemented through stochastic drop path augmentation \cite{huang2016deep}. As for width and expansion ratio, we follow Yu \emph{et al.}~\cite{yu2019universally} to construct switchable layers with shared weights but independent normalization layers, such that the corresponding layer can execute at different channel numbers from a predefined set, \emph{i.e.}, multiples of $16$ or $32$. 
Specifically, the expansion ratio $E_{i,j}$ is determined by the channels of the depth-wise $3\times3$ Conv in each \texttt{FFN}, and stage width $C_j$ is determined by aligning the output channels of the last projection ($1\times1$ Conv) of \texttt{FFN} and \texttt{MHSA} blocks. 
The switchable execution can be expressed as:
\begin{equation}
\small
    \hat{\mathbb{X}}_i = \gamma_c \cdot \frac{w^{:c} \cdot \mathbb{X}_i - \mu_c}{\sqrt{\sigma_c^2+\epsilon}} + \beta_c,
\end{equation}
where $w^{:c}$ refers to slicing the first $c$ filters of the weight matrix to obtain a subset of output, and $\gamma_c$, $\beta_c$, $\mu_c$, and $\sigma_c$ are the parameters and statistics of the normalization layer designated for width $c$. 
The supernet is pre-trained with Sandwich Rule~\cite{slimmable} by training the largest, the smallest, and randomly sampled two subnets at each iteration (we denote these subnets as max, min, rand-1, and rand-2 in Alg.~\ref{algorithm}).

\begin{algorithm}[]
\small
\caption{Evaluation-based search for size and speed}
\label{algorithm}
\begin{algorithmic}
\Require Latency lookup table $T : \{ \texttt{FFN}^{C,E}, \texttt{MHSA}^{C} \}$
\Ensure Subnet satisfying objectives: params, latency, or \texttt{MES}

\State $\rightarrow$ \textbf{Super-net Pretraining}: 
\For{epoch}
    \For{each iter}
        \For {subnet $\in$ \{min, rand-1, rand-2, max\} } 
            \State $\mathbb{Y} \gets \prod_i \{ \texttt{FFN}_i, \texttt{MHSA}_i \}(\mathbb{X}_i) $ 
            \State $\mathcal{L} \gets criterion(\mathbb{Y}, label)$, backpropagation
        \EndFor \Comment{Sandwich Rule}
        \State  Update parameters (AdamW~\cite{loshchilov2017decoupled})
    \EndFor 
\EndFor  \Comment{finish supernet training}
\State $\rightarrow$ \textbf{Joint search for size and speed: }
\State Initialize state  $S \gets \{ S_{N_{max}}, S_{C_{max}}, S_{E_{max}} \}$
\While{Objective not satisfied} 
\State Execute action $\hat{A} \gets {\arg\min}_A \frac{\Delta \texttt{Acc}}{\Delta \texttt{MES}}$
\State Update state frontier 
\EndWhile \Comment{get sub-net with target \texttt{MES}}

\State $\rightarrow$ \textbf{Train the searched architecture from scratch} 
\end{algorithmic}   
\end{algorithm}

\begin{table*}[ht]
\caption{{\textbf{Classification results on ImgeNet-1K.} 
We report the number of parameters, \emph{i.e.}, Params (M), GMACs, Training Epochs, and Top-1 accuracy for various methods.
The latency results are obtained by running models on iPhone 12 (Neural Engine) compiled with CoreMLTools, Pixel 6 (CPU) compiled with XNNPACK, and Nvidia A100 (GPU) compiled with TensorRT. The batch size is $1$ for models tested on iPhone 12 and Pixel 6, and $64$ for A100. (-) denotes unrevealed or unsupported models. $\dagger$ denotes we re-train the previous models with the exact same training recipe as our work. 
Different training seeds result in about 0.1\% fluctuation in accuracy. 
The latency is benchmarked with warmup and averaged over multiple runs, where the error ranges within 0.1 ms. 
}}
\label{tab:comparison}
\centering
\small
\scalebox{1.0}{
\begin{tabular}{lccccccccc}
\toprule
\multirow{2}{*}{Model}           & \multirow{2}{*}{Type}      & \multirow{2}{*}{Params (M)} & \multirow{2}{*}{GMACs} &  \multicolumn{3}{c}{Latency (ms)} & \multirow{2}{*}{$\texttt{MES} \uparrow$} & \multirow{2}{*}{Epochs} & \multirow{2}{*}{Top-1(\%)}  \\ \cline{5-7}
    &      &     &    &   iPhone 12  & Pixel 6  & A100     &   & &  \\
\hline
\hline
MobileNetV2$\times 1.0$     & CONV      &  3.5    & 0.3    & 0.9    & 25.3 &  5.0  & 102.9  &300& 71.8    \\
MobileNetV2$\times 1.0$ $\dagger$     & CONV      &  3.5    & 0.3    & 0.9    & 25.3 &  5.0  & 102.9  &300& 72.2    \\
MobileViT-XS       & Hybrid    &    2.3   & 0.7    & 7.3  & 64.4  & 11.7 & 15.6 &300& 74.8   \\
EdgeViT-XXS       & Hybrid    &   4.1    &  0.6   &   2.4    & 30.9 &  11.3 & 35.6 & 300  &  74.4   \\
Hydra Attention  &   Hybrid   & 3.5      &  0.38    & 4.1      & - &  9.7 & 22.6   & 300 & 75.6  \\
\rowcolor[gray]{0.92}
EfficientFormerV2-S0  &   Hybrid   & 3.5      &  0.40    & 0.9      & 20.8 & 6.6   & 102.9 & 300 / 450 & 75.7 / 76.2   \\
\hline
MobileNetV2$\times 1.4$     & CONV      &  6.1    & 0.6    & {1.2}     & 42.8 &  7.3 &  58.4  &300& 74.7    \\
MobileNetV2$\times 1.4$ $\dagger$     & CONV      &  6.1    & 0.6    & {1.2}     & 42.8 &  7.3 &  58.4  &300& 76.7    \\
MobileNetV3-L    & CONV      &  5.4    & 0.22   & 15.8    & - &  7.2 &  4.7  & 300 & 75.2    \\
FBNet-V3     & CONV      &  5.6    & 0.39   & 1.0    & - &  7.9 &  73.2  & 300 & 75.1    \\
EfficientNet-B0 & CONV      &  5.3  &  0.4   & 1.4            & 29.4 & 10.0  & 53.7  & 350 &   77.1  \\
DeiT-T          & Attention & 5.9       & 1.2      &    9.2  & 66.6 & 7.1    & 7.8 &300& 74.5    \\
EdgeViT-XS       & Hybrid    &   6.7    &  1.1   & 3.6  & 55.5 & 14.3 & 18.6 & 300 &   77.5  \\
LeViT-128S       & Hybrid    & 7.8      & 0.31     & 19.9    & 15.5 & 3.4 &  3.1  & 1000 & 76.6    \\
\rowcolor[gray]{0.92}
EfficientFormerV2-S1  &   Hybrid   & 6.1      &  0.65    & 1.1     & 33.3  &   8.8  &  63.8 & 300 / 450 & 79.0 / 79.7   \\
\hline
EfficientNet-B3 & CONV      & 12.0        & 1.8     &  5.3        & 123.8 & 35.0    & 9.4  & 350  & 81.6   \\
PoolFormer-s12  & Pool      &    12     & 2.0     &  1.5  & 82.4 & 14.5  & 33.3  & 300 & 77.2     \\
LeViT-192       & Hybrid    & 10.9      & 0.66     & 29.6 &30.1  &  5.2 &  1.8  & 1000 & 80.0    \\
MobileFormer-508M       & Hybrid    &    14.0   & 0.51    &  6.6  & 55.2 & 14.6  & 7.0 & 450 & 79.3    \\
UniNet-B1       & Hybrid    &    11.5   & 1.1    &  2.2     & 57.7  &  16.9& 23.2 & 300 & 80.8    \\
EdgeViT-S       & Hybrid    &   11.1    &   1.9  &   4.6   & 92.5 &  21.2 & 11.3 & 300 &  81.0   \\
EfficientFormer-L1       & Hybrid    &    12.3   & 1.3    & 1.4  & 50.7 & 8.4 & 35.3 & 300 & 79.2    \\
\rowcolor[gray]{0.92}
EfficientFormerV2-S2  &   Hybrid   &  12.6  &  1.25    & 1.6     & 57.2 &  14.5  & 30.5 & 300 / 450 & 81.6 / 82.0     \\
\hline
ResNet50        & CONV      &    25.5     & 4.1     & 2.5      &  167.5 & 9.0   & 13.7  &300& 78.5    \\
ResNet50 $\dagger$        & CONV      &    25.5     & 4.1     & 2.5      &  167.5 & 9.0   & 13.7  &300& 80.5    \\
ConvNext-T & CONV      &   29.0   &  4.5  &   83.7       & 340.5 &   28.8  & 0.4 & 300 & 82.1 \\
ResMLP-S24       & SMLP      &  30     & 6.0     & 7.6   & 325.4 & 17.4   &  4.2 &300& 79.4    \\
PoolFormer-s24  & Pool      &    21     & 3.6      &  2.4  & 154.3 & 28.2    & 15.7 & 300 & 80.3   \\
PoolFormer-s36  & Pool      &    31     & 5.2      &  3.5   & 224.9 & 41.2   &  8.9  & 300 & 81.4   \\
DeiT-S          & Attention & 22.5      & 4.5    &      11.8 & 218.2 & 15.5   &  3.1 &300& 81.2     \\
PVT-Small       & Attention &    24.5      & 3.8    &    24.4  & - & 23.8    &  1.4  &300& 79.8    \\
T2T-ViT-14      & Attention & 21.5      & 4.8     & -        & - & 21.0     & - &310& 81.5     \\
Swin-Tiny       & Attention &    29      & 4.5     & -       & - &22.0 & -  &300& 81.3    \\
CSwin-T       & Attention &    23      & 4.3       & -       & - & 28.7    & -  &300& 82.7  \\
LeViT-256       & Hybrid    & 18.9      & 1.12     & 31.4 & 50.7 & 6.7  &  1.3  & 1000 & 81.6    \\
LeViT-384       & Hybrid    & 39.1      & 2.35     & 48.8 & 102.2 & 10.2  &  0.6  & 1000 & 82.6    \\
Convmixer-768   & Hybrid    &  21.1 & 20.7  & 11.6  &  - & -  & 3.3 & 300 &    80.2   \\
NasViT-Supernet   & Hybrid    &  - &  1.9  &  -  &   - &  -  &  - &  360 &    82.9  \\
EfficientFormer-L3       & Hybrid    &    31.3   & 3.9    &   2.7  & 151.9 & 13.9  & 11.5 & 300 & 82.4    \\
EfficientFormer-L7      & Hybrid    &    82.1   & 10.2    &  6.6  & 392.9 & 30.7  & 2.9 & 300 & 83.3    \\
\rowcolor[gray]{0.92}
EfficientFormerV2-L  &   Hybrid   & 26.1  & 2.56 & 2.7     & 117.7 &   22.5 & 12.6  & 300 / 450 & 83.3 / 83.5    \\
\hline
\rowcolor[gray]{0.92}
Supernet  &   Hybrid   & 37.1  & 3.57 & 4.2     & - &   -  & 6.8  & 300 & 83.5    \\
\bottomrule
\end{tabular}
}
\end{table*}

\noindent\textbf{Discussion.} The pruning of our supernet is partially inspired by the slimmable network~\cite{slimmable}. However, the differences are also significant. First, the search objective is different. We apply the introduced joint search objective for optimizing model size and efficiency (Sec.~\ref{sec:objective}).
Second, the search actions are different. Depth is pruned through the reduction of each \emph{block}, which is possible since we unify the design and only adopt two blocks: Feed Forward Network (Sec.~\ref{sec:mixer}) and attention block. The way of pruning the depth is different from the slimmable network.
Unifying all the flexible search actions under one joint objective has not been studied for transformers before. 

\subsubsection{Search Algorithm}
Now that search objective, search space, and supernet are formulated, we present the search algorithm.
Since the supernet is executable at elastic depth and switchable width, we can search the subnetworks with the best Pareto curve by analyzing the efficiency gain and accuracy drop with respect to each slimming action. 
We define the action pool as: 
\begin{equation}
\small
    A \in \{ A_{N[i,j]}, A_{C[j]}, A_{E[i,j]} \},
\end{equation}
where $A_{N[i,j]}$ denotes slimming each block, $A_{C[j]}$ refers to shrinking the width of a stage, and $A_{E[i,j]}$ denotes slimming each \texttt{FFN} to a smaller expansion. 
Initializing the state with full depth and width (largest subnet), we evaluate the accuracy outcome ($\Delta \texttt{Acc}$) of each frontier action on a validation partition of ImageNet-1K, which only takes about 4 GPU-minutes. 
Meanwhile, parameter reduction ($\Delta \texttt{Params}$) can be directly calculated from layer properties, \emph{i.e.}, kernel size, in-channels, and out-channels. 
We obtain the latency reduction ($\Delta \texttt{Latency}$) through a pre-built latency look-up table measured on iPhone 12 with CoreMLTools. 
With the metrics in hand, we can compute $\Delta \texttt{MES}$ through $\Delta \texttt{Params}$ and $\Delta \texttt{Latency}$, and choose the action with the minimum per-\texttt{MES} accuracy drop: $\hat{A} \gets {\arg\min}_A \frac{\Delta \texttt{Acc}}{\Delta \texttt{MES}}$.
It is noteworthy that though the action combination is enormous, we only need to evaluate the frontier one at each step, which is linear in complexity. Details can be found in Alg.~\ref{algorithm}.

\section{Experiments}

\begin{table*}[]
    \centering
      \caption{
\textbf{Object detection \&
instance segmentation} on MS COCO 2017 with the Mask RCNN pipeline.}
      \small
      \resizebox{1\linewidth}{!}{
        \begin{tabular}{l|c|cccccc|c}
\toprule
\multirow{2}{*}{Backbone} & \multirow{2}{*}{Params (M)} & \multicolumn{6}{c}{Detection \& Instance Segmentation} & \multicolumn{1}{c}{Semantic}  \\ \cline{3-9}
                          &                         & AP$^{box}$    & AP$^{box}_{50}$   & AP$^{box}_{75}$   & AP$^{mask}$    & AP$^{mask}_{50}$   & AP$^{mask}_{75}$   & mIoU   \\
                          \hline
                          \hline
ResNet18                  &     11.7        & 34.0   & 54.0    & 36.7    & 31.2   & 51.0    & 32.7   & 32.9   \\
PoolFormer-S12            &     12.0          & 37.3   & 59.0    & 40.1    & 34.6   & 55.8    & 36.9     & 37.2  \\
EfficientFormer-L1        &     12.3              & 37.9   & 60.3    & 41.0    & 35.4   & 57.3    & 37.3   &  38.9  \\
\rowcolor[gray]{0.92}
EfficientFormerV2-S2     &     12.6              &  43.4  &  65.4   & 47.5  &    39.5    &   62.4      &  42.2   & 42.4  \\
\hline
ResNet50                  &      25.5           & 38.0   & 58.6    & 41.4    & 34.4   & 55.1    & 36.7  &  36.7  \\
PoolFormer-S24            &     21.0            & 40.1   & 62.2    & 43.4    & 37.0   & 59.1    & 39.6  &  40.3 \\
Swin-T                    &     29.0              & 42.2   & 64.4    & 46.2    & 39.1   & 64.6    & 42.0   & 41.5  \\
EfficientFormer-L3        &         31.3     & 41.4   & 63.9    & 44.7    & 38.1   & 61.0    & 40.4   &  43.5 \\
\rowcolor[gray]{0.92}
EfficientFormerV2-L      &       26.1      & 44.7   &  66.3  &  48.8   &  40.4   &  63.5   & 43.2     &   45.2   \\
\bottomrule
\end{tabular}}\label{tab:coco}
\end{table*}


\subsection{ImageNet-1K Classification}
\noindent\textbf{Implementation Details.}
We implement the model through PyTorch 1.12~\cite{paszke2019pytorch} and Timm library~\cite{rw2019timm}, and use $16$ NVIDIA A100 GPUs to train our models.
We train the models from scratch by $300$ and $450$ epochs on ImageNet-1K~\cite{deng2009imagenet}, with AdamW~\cite{loshchilov2017decoupled} optimizer. 
Learning rate is set to $10^{-3}$ per $1,024$ batch size with cosine decay. 
We use a standard image resolution, \emph{i.e.}, $224\times224$, for both training and testing. 
Similar to DeiT~\cite{touvron2021training}, we use RegNetY-16GF~\cite{radosavovic2020designing} with $82.9\%$ top-1 accuracy as the teacher model for hard distillation.  
We use three testbeds to benchmark the latency:
\begin{itemize}[leftmargin=1em]
    \item \textbf{iPhone 12 - NPU.} We get the latency on iPhone 12 (iOS 16) by running the models on Neural Engine (NPU). The models (batch size of $1$) are compiled with CoreML~\cite{coreml2021}.
    \item \textbf{Pixel 6 - CPU.} We test model latency on Pixel 6 (Android) CPU. To obtain the latency for most works under comparison, we replace the activation from \emph{all} models to ReLU to get fair comparisons. The models (batch size of $1$) are compiled with XNNPACK~\cite{XNNPACK}.
    \item \textbf{Nvidia GPU.} We also provide the latency on a high-end GPU--Nvidia A100. The models (batch size of $64$) are deployed in ONNX~\cite{ONNX} and executed by TensorRT~\cite{TensorRT}. 
\end{itemize}
\noindent\textbf{Evaluation on Single Metric.}
We show the comparison results in Tab.~\ref{tab:comparison}, which includes \emph{the most recent and representative} works on vision transformers and CNNs. The works that \emph{do not have public models} or \emph{are not compatible with mobile devices}~\cite{cai2022efficientvit,deformableattention,liu2021swin2} are not contained in Tab.~\ref{tab:comparison}.
EfficientFormerV2 series achieve the SOTA results on a single metric, \emph{i.e.}, number of parameters or latency. 
For model size, EfficientFormerV2-S0 outperforms EdgeViT-XXS~\cite{pan2022edgevits} by $1.3\%$ top-1 accuracy with even $0.6$M fewer parameters and MobileNetV2$\times1.0$~\cite{sandler2018mobilenetv2} by $3.5\%$ top-1 with similar number of parameters. For large models, EfficientFormerV2-L model achieves identical accuracy to recent EfficientFormer-L7~\cite{efficientformer} while being $3.1\times$ smaller. 
As for speed, with comparable or lower latency, EfficientFormerV2-S2 outperforms UniNet-B1~\cite{liu2022uninet}, EdgeViT-S~\cite{pan2022edgevits}, and EfficientFormer-L1~\cite{efficientformer} by $0.8\%$, $0.6\%$ and $2.4\%$ top-1 accuracy, respectively. 
We hope the results can provide practical insight to inspire future architecture design: \emph{modern deep neural networks are robust to architecture permutation, optimizing the architecture with joint constraints, such as latency and model size, will not harm individual metrics.}

\noindent\textbf{Jointly Optimized Size and Speed.}
Further, we demonstrate the superior performance of EfficientFormerV2 when considering both model size and speed. 
Here we use \texttt{MES} as a more practical metric to assess mobile efficiency than using size or latency alone.
EfficientFormerV2-S1 outperforms MobileViT-XS~\cite{mehta2021mobilevit}, EdgeViT-XXS~\cite{pan2022edgevits}, and EdgeViT-XS~\cite{pan2022edgevits} by $4.2\%$, $4.6\%$, and $1.5\%$ top-1, respectively, with 
far higher \texttt{MES}. With $1.8\times$ higher \texttt{MES},  EfficientFormerV2-L outperforms MobileFormer-508M~\cite{chen2021mobile} by $4.0\%$ top-1 accuracy. 
The evaluation results answer the central question raised at the beginning: \emph{with the proposed mobile efficiency benchmark (Sec.~\ref{sec:objective}), we can avoid entering a pitfall achieving seemingly good performance on one metric while sacrificing too much for others. Instead, we can obtain efficient mobile ViT backbones that are both light and fast. }

\subsection{Downstream Tasks}

\begin{table}[]
\small
    \centering
        \caption{
\textbf{Ablation analysis of search algorithms.} Our proposed fine-grained search with joint constraints on size and speed achieves better results than random search and the coarse-grained, single objective search from EfficientFormer~\cite{efficientformer}. 
Latency is measured on iPhone 12.
}
\resizebox{1.0\linewidth}{!}{
\begin{tabular}{l|c|c|c}
\toprule
Search Algorithm & Params (M)& Latency (ms) & Top-1 ($\%$) \\ \hline
Random 1    &    3.5    &   1.0   &   74.7 \\
Random 2    &    3.5    &   1.0   &    75.0  \\
\rowcolor[gray]{0.92}
EfficientFormerV2 (Ours) &    3.5    &   0.9      &   75.7   \\
\hline
EfficientFormer~\cite{efficientformer} & 3.1 & 0.9 & 74.2 \\
\rowcolor[gray]{0.92}
EfficientFormerV2 (Ours) &3.1 & 0.9 & 75.0 \\ 
\bottomrule
\end{tabular}}\label{tab:ablate_search}
\end{table}


\noindent\textbf{Object Detection and Instance Segmentation.}  We apply EfficientFormerV2 as backbone in Mask-RCNN~\cite{he2017mask} pipeline and experiment on MS COCO 2017~\cite{lin2014microsoft}. The model is initialized with ImageNet-1K pretrained weights. We use AdamW~\cite{loshchilov2017decoupled} optimizer with an initial learning rate as $2\times10^{-4}$ and conduct training for $12$ epochs with resolution as $1333\times 800$. Following Li \emph{et al.}~\cite{li2022next}, we apply a weight decay as $0.05$ and freeze the normalization layers in the backbone.
As in Tab.~\ref{tab:coco}, with similar model size, our EfficientFormerV2-S2 outperform PoolFormer-S12~\cite{yu2021metaformer} by $6.1$ AP$^{box}$ and $4.9$ AP$^{mask}$. EfficientFormerV2-L outperforms EfficientFormer-L3~\cite{efficientformer} by $3.3$ AP$^{box}$ and $2.3$ AP$^{mask}$. 

\noindent\textbf{Semantic Segmentation.}  We perform experiments on ADE20K \cite{zhou2017scene}, a challenging scene segmentation dataset with $150$ categories. 
Our model is integrated as a feature encoder in Semantic FPN~\cite{kirillov2019panoptic} pipeline, with ImageNet-1K pretrained weights. 
We train our model on ADE20K for $40$K iterations with batch size as $32$ and learning rate as $2\times 10^{-4}$ with a poly decay by the power of $0.9$. 
We apply weight decay as $10^{-4}$ and freeze the normalization layers.
Training resolution is $512\times 512$, and we employ a single scale testing on the validation set. 
As in Tab.~\ref{tab:coco}, EfficientFormerV2-S2 outperforms PoolFormer-S12 \cite{yu2021metaformer} and EfficientFormer-L1 \cite{efficientformer} by $5.2$ and $3.5$ mIoU, respectively.

\subsection{Ablation Analysis on Search Algorithm}

We compare the proposed search algorithm with the random search and the one from EfficientFormer~\cite{efficientformer}. As seen in Tab.~\ref{tab:ablate_search}, our search algorithm obtains models with much better performance than random search, \emph{i.e.}, Random 1 and Random 2. Compared with EfficientFormer~\cite{efficientformer}, we achieve higher accuracy under similar parameters and latency, demonstrating the effectiveness of fine-grained search and joint optimization of latency and size. 


\section{Discussion and Conclusion}
In this work, we comprehensively study transformer backbones, identify inefficient designs, and introduce mobile-friendly novel architectural changes.
We further propose a fine-grained joint search on size and speed and obtain the EfficientFormerV2 model family. We extensively benchmark and compare our work with existing studies on different hardware and demonstrate that EfficientFormerV2 is both lightweight, ultra-fast in inference speed and high performance. 
Since we focus on size and speed,
one future direction is to apply the joint optimization methodology to subsequent research exploring other critical metrics, such as memory footprint and CO$_2$ emission.

\section{Acknowledgements}
This work is partly supported by the Army Research Office/Army Research Laboratory via grant W911-NF-20-1-0167 to Northeastern University, \and CNS1909172.

{\small
\bibliographystyle{ieee_fullname}
\bibliography{egbib}
}

\newpage
\appendix
\section{More Experimental Details and Results}
\noindent\textbf{Training hyper-parameters.}
We provide the detailed training hyper-parameters for the ImageNet-1K~\cite{deng2009imagenet} classification task in Tab.~\ref{tab:hyperparam}, which is a similar recipe following DeiT~\cite{touvron2022deit}, LeViT~\cite{levit}, and EfficientFormer~\cite{efficientformer} for fair comparisons.


\begin{table}[h]
\caption{Training hyper-parameters for ImageNet-1K classification task. The drop path rate is for the [S0, S1, S2, L] model series. }
\label{tab:hyperparam}
\centering
\begin{tabular}{cc}
\toprule
Hyperparameters & Config                \\
\hline
optimizer       & AdamW                 \\
learning rate   & $0.001\times$(BS/$1024$)     \\
LR schedule     & cosine                \\
warmup epochs   & $5$                     \\
training epochs & $300$                   \\
weight decay    & $0.025$                 \\
augmentation    & RandAug($9$, $0.5$)       \\
color jitter    & $0.4$                   \\
gradient clip   & $0.01$                  \\
random erase    & $0.25$                  \\
label smooth     & $0.1$                   \\
mixup           & $0.8$                   \\
cutmix          & $1.0$                   \\
drop path       & {[}$0, 0, 0.02, 0.1${]} \\
\bottomrule
\end{tabular}
\end{table}

\noindent\textbf{Results without distillation.}
We provide our models trained \emph{without} distillation in Table.~\ref{tab:wodistill}.
Compared with representative works trained without distillation, \emph{e.g.}, MobileNetV2 \cite{sandler2018mobilenetv2}, MobileFormer \cite{chen2022mobile} (trained with longer epochs), EdgeViT \cite{pan2022edgevits}, and PoolFormer \cite{yu2021metaformer}, our models still achieve better latency-accuracy trade-offs. 
\begin{table}[]
\small
\centering
\caption{Results without distillation. }
\scalebox{0.65}{
\begin{tabular}{cccccc}
\hline
Model                & Params (M) & MACs (G) & Latency (ms) & Epochs & Top-1 (\%) \\
\hline
MobileNetV2 1.0      & 3.5        & 0.30     & 0.9          & 300    & 71.8       \\
\rowcolor[gray]{0.92}
EfficientFormerV2-S0 & 3.5        & 0.40     & 0.9          & 300    & 73.7       \\ 
\hline
EdgeViT-XS           & 6.7        & 1.10     & 3.6          & 300    & 77.5       \\
\rowcolor[gray]{0.92}
EfficientFormerV2-S1 & 6.1        & 0.65     & 1.1          & 300    & 77.9       \\
\hline
MobileFormer-508M    & 14.0       & 0.51     & 6.6          & 450    & 79.3       \\
PoolFormer-s12       & 12.0       & 2.0      & 1.5          & 300    & 77.2       \\
\rowcolor[gray]{0.92}
EfficientFormerV2-S2 & 12.6       & 1.25     & 1.6          & 300    & 80.4      \\
\hline
\end{tabular}
}
\label{tab:wodistill}
\end{table}

\noindent\textbf{Analysis on attention bias.}

\begin{table}[]

\caption{Analysis of explicit position encoding (Attention Bias). We use EfficientFormerV2-S1 for the experiments.}
\label{tab:long_epoch}
\centering
\begin{tabular}{cccc}
\toprule
 Params (M) & Epoch & Attention Bias & Top-1 (\%) \\
\hline
6.10       & 300   & Y    & 79.0       \\
6.08       & 300   & N    & 78.8       \\ \hline
6.10       & 450   & Y    & 79.7       \\
6.08       & 450   & N    & 79.5      \\
\bottomrule
\end{tabular}
\end{table}



Attention Bias is employed to serve as explicit position encoding. 
On the downside, attention bias is resolution sensitive, making the model fragile when migrating to downstream tasks. 
By deleting attention bias, we observe $0.2\%$ drop in accuracy for both $300$ and $450$ training epochs (Attention Bias as Y \emph{vs.} N in Tab.~\ref{tab:long_epoch}), showing that EfficientFormerV2 can still preserve a reasonable accuracy without explicit position encoding. 



\section{More Ablation Analysis of Search Algorithm}

\noindent\textbf{Importance of Expansion Ratios.}
We first discuss the necessity to search for expansion ratios on top of width. 
As in Tab.~\ref{tab:ablate_expansion}, we show that, by adjusting width to maintain an identical budget, \emph{i.e.}, the same number of parameters for each model, varying the expansion ratio incurs considerable difference in performance. 
As a result, we can not obtain Pareto optimality by solely searching for width while setting a fixed expansion ratio. 

\begin{table}[h]
\caption{Ablation analysis on expansion ratios. Varying expansion ratios lead to different results even with the same number of parameters. Latency is obtained on iPhone 12.}
\label{tab:ablate_expansion}
\centering
\resizebox{1.0\linewidth}{!}{
\begin{tabular}{cccc}
\toprule
Expansion ratio & Params (M) & Latency (ms) & Top-1 (\%) \\
\hline
4         & 13.4       & 1.6          & 81.8       \\
2         & 13.4       & 1.6          & 81.6       \\
1         & 13.4       & 1.6          & 81.1      \\
\bottomrule
\end{tabular}}
\end{table}

\noindent\textbf{Analysis of Searching the Expansion Ratios.}
We verify the performance of different search algorithms in Tab.~\ref{tab:ablate_CE_search}. 
We obtain the baseline result using the search pipeline in EfficientFormer~\cite{efficientformer} to search only for the depth and width. 
With a budget of $7$M parameters, we obtain a subnetwork with $79.2\%$ top-1 accuracy on ImageNet-1K. 
Then, we apply a simple magnitude-based pruning to determine expansion ratios in a fine-grained manner. Unfortunately, the performance is not improved.
Though searching for expansion ratios is important (Tab.~\ref{tab:ablate_expansion}), it is non-trivial to achieve Pareto optimality with simple heuristics. 
Finally, we apply our fine-grained search method and obtain a subnetwork with $79.4\%$ top-1 accuracy, demonstrating the effectiveness of our approach. 

\begin{table}[h]
\caption{Ablation on search methods for depth, width, and expansion ratios. EfficientFormer~\cite{efficientformer} merely searches for depth and width. On top of EfficientFormer~\cite{efficientformer}, we perform network pruning to decide channel numbers for stage width and expansion ratios. Finally, we show the results of our search algorithm for jointly optimizing depth, width, and expansions. }
\label{tab:ablate_CE_search}
\centering
\resizebox{1.0\linewidth}{!}{
\begin{tabular}{cccc}
\toprule
Method  & Params (M) & Latency (ms) & Top-1 (\%) \\
\hline
From EfficientFormer~\cite{efficientformer}   & 7.0        & 1.15         & 79.2       \\
From EfficientFormer~\cite{efficientformer} + Pruning & 7.0        & 1.15         & 79.2       \\
\rowcolor[gray]{0.92}
Ours    & 7.0        & 1.15         & 79.4       \\
\bottomrule
\end{tabular}}
\end{table}

\noindent\textbf{Analysis of Different $\alpha_{latency}$ and $\alpha_{size}$ in Eqn. 1.} Here, we provide the results for analyzing how different values of $\alpha_{latency}$ and $\alpha_{size}$ can impact the search results in Tab.~\ref{tab:search_alpha}.
Our search algorithm is stable to different $\alpha$ settings. 
Increasing the weight of size ($\alpha_{size}$) leads to slower models. Our current setting ($\alpha_{latency}$ as $1.0$ and $\alpha_{size}$ as $0.5$) is determined by aligning with recent works, 
\emph{e.g.,} EdgeViT, UniNet, etc, to make fair comparisons.

\begin{table}[h]
\centering
\caption{Analysis of the $\alpha_{latency}$ and $\alpha_{size}$ in search algorithm.}\label{tab:search_alpha}
\resizebox{1\linewidth}{!}{
\begin{tabular}{ccccc}
\toprule
$\alpha_{latency}$ & $\alpha_{size}$ & Params (M) & Latency & Top-1 (\%) \\
\hline
1.0       & 1.0    &    3.5    &   1.1 ms     &    77.0   \\
0.5       & 1.0    &    3.5    &  1.3 ms      &  77.3     \\
\hline
\rowcolor[gray]{0.92}
1.0       & 0.5    & 3.5    & 0.9 ms    & 75.7  \\
\bottomrule
\end{tabular}
}
\end{table}

\begin{figure}[h]
    \centering
    \includegraphics[width=1\linewidth]{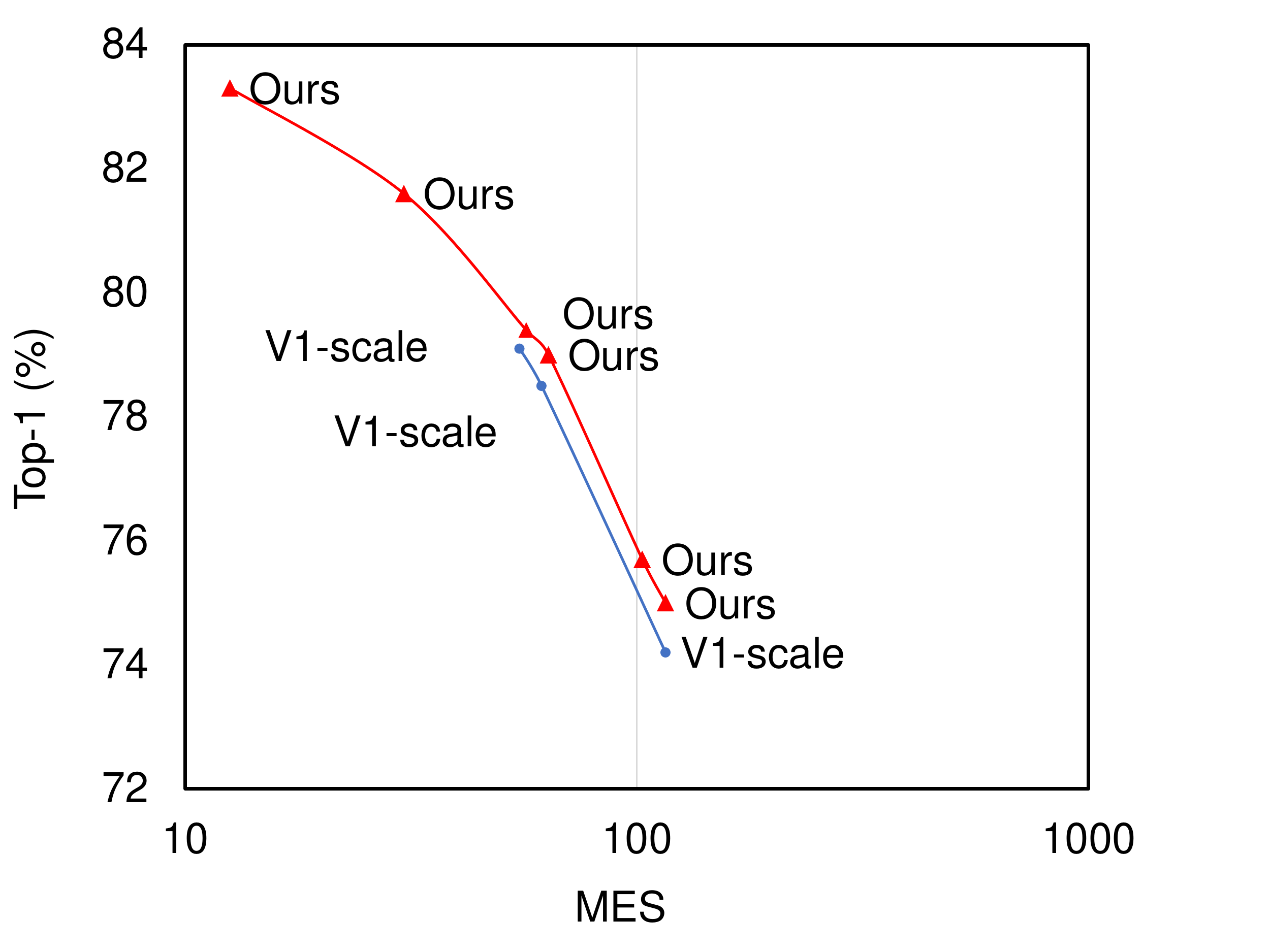}  
    \caption{Comparisons between our search method (Ours) and the search pipeline from EfficientFormer~\cite{efficientformer} (denoted as V1-scale), starting from the same supernet trained on ImageNet-1K. }
    \label{fig:compare_search}
\end{figure}

\noindent\textbf{Visualization of Search Results.}
In Fig.~\ref{fig:compare_search}, we visualize the performance of the searched subnetworks, including the networks obtained by using the search algorithm from EfficinetFormer\cite{efficientformer} and networks found by our fine-grained joint search. 
We employ \texttt{MES} as an efficiency measurement and plot in logarithmic scale. The results demonstrate the advantageous performance of our proposed search method.

\noindent\textbf{Design Choice Ablation.}
We ablate our network design choices on detection/instance segmentation task,
and prove that the conclusions from ImageNet-1K classification task can  \emph{transfer}.
We train EfficientFormerV2-S2 on \emph{MS-COCO} dataset from \textbf{scratch} for $12$ epochs \emph{without} ImageNet pretraining. 
The results are included in the Tab.~\ref{tab:downstream}. 
For instance, Sec.3.1 refers to falling back to DWConv mixer instead of FFN. 
Our design holds clear advantages. 
In addition,
without our proposed stride attention (Sec.3.4), the model encounters memory issues and can not run on mobile. Note that Sec.3.2 is not included as 5 stage network is not a common practice in detection tasks. 
\begin{table}[]
\small
\centering
\caption{Generalization of design choices on detection and instance segmentation. Configuration matches Tab.1 in paper. For instance, Sec.3.1 refers to falling back to DWConv mixer instead of FFN. Without our proposed stride attention (Sec.3.4), the model encounters memory issues and cannot run on mobile. Note that Sec.3.2 is not included as 5 stage network is not a common practice in detection tasks. }
\begin{tabular}{cccc}
\hline
Configuration        & Latency (ms) & $\text{AP}^{box}$ & $\text{AP}^{mask}$ \\
\hline
EfficientFormerV2-S2 & 187.9        & 33.5    & 31.2     \\
\hline
Sec.3.1              & 181.1        & 31.6    & 29.5     \\
Sec.3.3              & 187.2        & 33.4    & 31.2     \\
Sec.3.4              & Failed       & 33.9    & 31.6     \\
Sec.3.5              & 187.7        & 32.7    & 30.6     \\
\hline
\end{tabular}
\label{tab:downstream}
\end{table}

\noindent\textbf{More random models and the cost analysis of searching via supernet \emph{vs.} random. }
We sample more random models ($10$) with a more extensive latency range to compare against our searched models. 
\begin{figure}
    \centering
\includegraphics[trim={1mm 0mm 1mm 8mm},clip,width=0.8\linewidth]{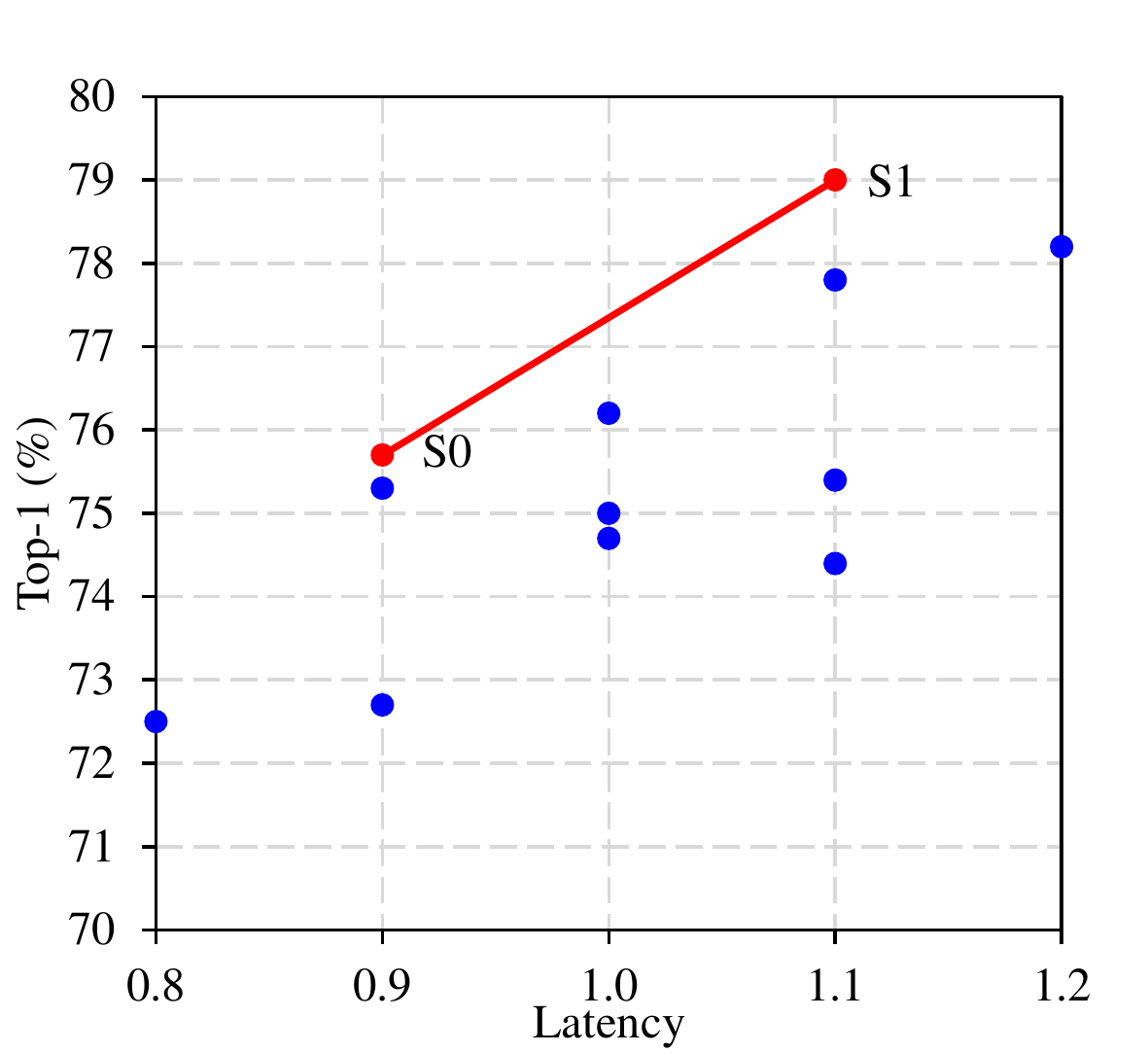}
  \caption{Comparisons with more random sampled models. We take 0.1ms as the significant digit based on mobile measurement precision. }
\label{fig:more_random}
\end{figure} 
As seen in Fig.~\ref{fig:more_random}, searching models by our approach (red line) gets better performance than random search (blue dots). 
Our supernet training takes $37$ GPU days (A100), which is $4.6\times$ the training time of the L model ($8$ GPU days). 
However, assuming at least $10$ random subnets are needed to search each candidate, the cost of random search for L-level model itself accumulates to $80$ GPU days ($2\times$ longer than supernet). 
Also, the cost of random search further scales up for multiple networks (four in our work).
Thus, our search method is more efficient than random search.

\begin{figure}
    \centering
\includegraphics[trim={1mm 0mm 1mm 6mm},clip,width=0.8\linewidth]{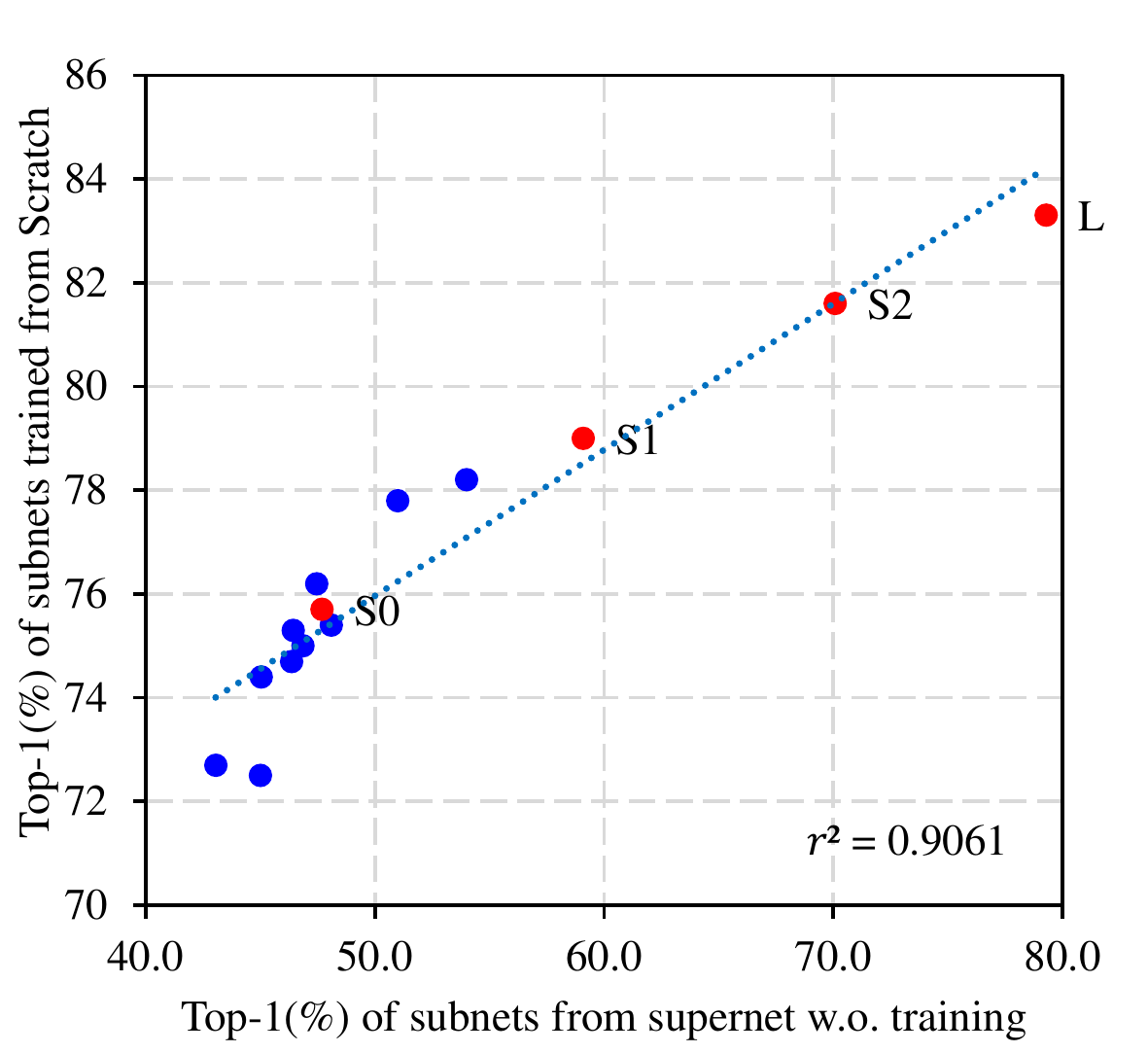}
\caption{Subnet evaluation vs. training from scratch. We report our searched models (red) along with random ones (blue). }
    \label{fig:correlation}
\end{figure}

\noindent \textbf{Accuracy of subnets from supernet and their correlation to final accuracy.} 
In the Figure.~\ref{fig:correlation},
we show the accuracy of multiple subnets obtained from the supernet and their correlation to final accuracy (training from scratch).
\noindent We refer EagleEye \cite{li2020eagleeye} for comparison. 
Through effectively-trained supernet, we obtain higher subnet evaluation accuracy ($>40\%$, \emph{v.s.} $<10\%$ in EagleEye), as well as better correlations to final accuracy ($r^2=0.91$, \emph{v.s} $0.63$ in EagleEye) measured by Pearson correlation coefficient.

\section{Network Configurations}
The detailed network architectures for EfficientFormerV2-S0, S1, S2, and L are provided in Tab.~\ref{tab:achitecture}. 
We report the stage resolution, width, depth, and per-block expansion ratios. 
\begin{table*}[ht]
\caption{Architecture details of EfficientFormerV2.}
\label{tab:achitecture}
\begin{tabular}{c|c|c|c|c|c|c|c}
\toprule
\multirow{2}{*}{Stage} & \multirow{2}{*}{Resolution} & \multirow{2}{*}{Type} & \multirow{2}{*}{Config} & \multicolumn{4}{c}{EfficientFormerV2}                                   \\ \cline{5-8}
                       &                             &                       &                         & S0                & S1                & S2            & L               \\
                       \hline
\multirow{4}{*}{stem}  & \multirow{2}{*}{$\frac{H}{2}\times \frac{W}{2}$}         & \multirow{2}{*}{\texttt{Conv}} & Kernel, Stride           & $3\times 3$, $2$            & $3\times 3$, $2$            & $3\times 3$, $2$        & $3\times 3$, $2$          \\ \cline{4-8}
                       &                             &                       & N, C                    & $1$, $16$             & $1$, $16$             & $1$, $16$         & $1$, $20$           \\ \cline{2-8}
                       & \multirow{2}{*}{$\frac{H}{4}\times \frac{W}{4}$}         & \multirow{2}{*}{\texttt{Conv}} & Kernel, Stride           & $3\times 3$, $2$            & $3\times 3$, $2$            & $3\times 3$, $2$        & $3\times 3$, $2$          \\ \cline{4-8}
                       &                             &                       & N, C                    & $1$, $32$             & $1$, $32$             & $1$, $32$         & $1$, $40$           \\
                       \hline
\multirow{2}{*}{1}     & \multirow{2}{*}{$\frac{H}{4}\times \frac{W}{4}$}         & \multirow{2}{*}{\texttt{FFN}}  & N, C                    & $2$, $32$             & $3$, $32$             & $4$, $32$         & $5$, $40$           \\ \cline{4-8}
                       &                             &                       & E                       & {[}$4,4${]}         & {[}$4,4,4${]}       & {[}$4,4,4,4${]} & {[}$4,4,4,4,4${]} \\
                       \hline
\multirow{2}{*}{2}     & \multirow{2}{*}{$\frac{H}{8}\times \frac{W}{8}$}         & \multirow{2}{*}{\texttt{FFN}}  & N, C                    & $2, 48$             & $3, 48$             & $4, 64$         & $5, 80$           \\ \cline{4-8}
                       &                             &                       & E                       & {[}$4,4${]}         & {[}$4,4,4${]}       & {[}$4,4,4,4${]} & {[}$4,4,4,4,4${]} \\
                       \hline
\multirow{3}{*}{3}     & \multirow{3}{*}{$\frac{H}{16}\times \frac{W}{16}$}        & \multirow{2}{*}{\texttt{FFN}}  & N, C                    & $6, 96$             & $9, 120$            & $12, 144$       & $15, 192$         \\  \cline{4-8}
                       &                             &                       & E                       & {[}$4,3,3,3,4,4${]} & {[}$4(\times5),3(\times4)${]}     & {[}$4(\times6),3(\times6)${]} & {[}$4(\times8),3(\times7)${]}   \\
                      \cline{3-8}
                       &                             & \texttt{MHSA}                  & N                       & $2$                 & $2$                 & $4$             & $6$               \\
                       \hline
\multirow{3}{*}{4}     & \multirow{3}{*}{$\frac{H}{32}\times \frac{W}{32}$}        & \multirow{2}{*}{\texttt{FFN}}  & N, C                    & $4, 176$            & $6, 224$            & $8, 288$        & $10, 384$         \\ \cline{4-8}
                       &                             &                       & E                       & {[}$4,3,3,4${]}     & {[}$4,4,3,3,4,4${]} & {[}$4(\times4),3(\times4)${]} & {[}$4(\times6), 3(\times4)${]}  \\
                       \cline{3-8}
                       &                             & \texttt{MHSA}                  & N                       & $2$                 & $2$                 & $4$             & $6$  \\
                       \bottomrule
\end{tabular}
\end{table*}

\end{document}